\documentclass[conference, review, onecolumn]{IEEEtran}

\usepackage{fancyhdr}

\usepackage{subfigure}
\usepackage{graphicx}

\usepackage{indentfirst}
\usepackage{latexsym}

\usepackage{dsfont}
\usepackage{txfonts}
\usepackage{amsfonts}

\usepackage{amssymb}

\begin{document}
\title{CRH: A Simple Benchmark Approach to Continuous Hashing}

\author{\IEEEauthorblockN{Miao Cheng}  
\IEEEauthorblockA{College of Information Engineering\\
Qingdao University\\
Qingdao, China\\
E-mail: mcheng@qdu.edu.cn\\}
\and
\IEEEauthorblockN{Ah Chung Tsoi}   
\IEEEauthorblockA{Faculty of Information Technology\\
Macau University of Science and Technology\\
Macau S.A.R., China\\
E-mail: actsoi@must.edu.mo}}


%


\maketitle
\thispagestyle{plain}

\fancypagestyle{plain}{
\fancyhf{}	
\fancyfoot[L]{}
\fancyfoot[C]{}
\fancyfoot[R]{}
\renewcommand{\headrulewidth}{0pt}
\renewcommand{\footrulewidth}{0pt}
}

\pagestyle{fancy}{
\fancyhf{}
\fancyfoot[R]{}}
\renewcommand{\headrulewidth}{0pt}
\renewcommand{\footrulewidth}{0pt}

\begin{abstract}
In recent years, the distinctive advancement of handling huge data promotes the evolution of ubiquitous computing and analysis technologies. With the constantly upward system burden and computational complexity, adaptive coding has been a fascinating topic for pattern analysis, with outstanding performance. In this work, a continuous hashing method, termed continuous random hashing (CRH), is proposed to encode sequential data stream, while ignorance of previously hashing knowledge is possible. Instead, a random selection idea is adopted to adaptively approximate the differential encoding patterns of data stream, e.g., streaming media, and iteration is avoided for stepwise learning. Experimental results demonstrate our method is able to provide outstanding performance, as a benchmark approach to continuous hashing.
\end{abstract}


%
\IEEEpeerreviewmaketitle

\section{Introduction and Related Work}
Similarity search, as the basic issue of most information systems, addresses the problem of learning the similar data from buckets with respect to a given query data point \cite{LSHpSD04Mayur}\cite{LSH08Alexandr}. It is also known as \emph{Approximate Nearest Neighbour} (ANN) search \cite{Jegou11PQ} in large data sets with an identical demand. Generally, it is required to learn the compact binary codes of data, and then retrieval and recall of information can be implemented with codes matching \cite{Norouzi11MLH}.

The basic idea has resorted to use several hash functions mapping a bridge between input features and data codes. Then, the obtained short binary strings act as an index to directly access elements with comparison. Typical examples include Locality-Sensitive Hashing (LSH) \cite{Charikar02SER}, which works by projecting the data points to hyper planes and then obtaining the hash codes by certain bounding of thresholds, e.g.,
\begin{equation}
h_k \left( {f_k \left( {{x}} \right)} \right) = \left\{
\begin{array}{cc}
{1,} & if ~ {f_k \left( {x} \right) \ge \varepsilon }\\
{0,} & {otherwise.}
\end{array}
\right.
\end{equation}
where $ \varepsilon $ indicates the given threshold for binary coding. Furthermore, the similarities corresponding to different data points, e.g., $ x_i $ and $ x_j $, are expected to be preserved in the hashing codes $ h \left( f_k \left( x_i \right) \right) $ and $ h \left( f_k \left( x_j \right) \right) $ , and ANN indexing is available for extended feedback search. And a data mapping function $ f \left( x_i \right) $ is usually considered, which refers to linear transformations $ w^T{x_i} + {b} $. Here, $ w $ is a projection direction and $ b $ is the mapping offset.
The similar idea has been widely adopted to design variants of the hash functions with simple projections,  while keeping the main patterns absorbed into binary hash codes. The outstanding advantage over others is their learning are mostly data-independent, and good compatibility is able to be provided for wide applications \cite{Salakhutdinov09SH}\cite{Torralba08SC}\cite{Strecha12LDAhash}.

\begin{figure*}
  \centering
  \includegraphics[width=.56\textwidth]{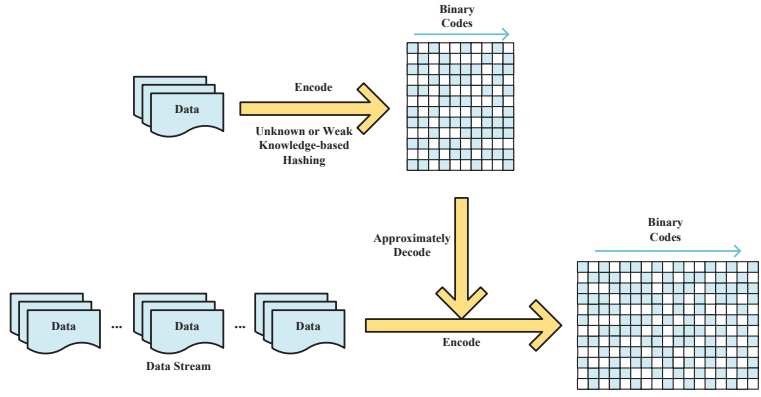}
  \caption{The basic idea of continuous random hashing.}
\end{figure*}
Another quite typical attempt is to define an objective function, usually a hash function as optimized encoder \cite{Norouzi11MLH}\cite{Carreira15HBA}, so that binary codes can be obtained with continuous relaxation. In light of Laplacian eigenmaps, Spectral Hashing (SH) \cite{Weiss09SH} computes approximate solutions by relaxing the binary constraints.
Based on this idea, some variations use the codes from spectral hashing to define the hash function as labels of a binary classifier. To hash the similar data among a large data set, a common approach is used to query data patterns with similarity preserving. It has been a common view that iterative optimization is applied to find out the best solution by defining objective functions \cite{Gong11ITQ}\cite{Norouzi11MLH}. Nevertheless, it requires more calculations to meet the local optimum usually. And it would be ideal for complicated applications if few steps are adopted in encoder of binary hashing.

Until now, many outlets seek for data patterns, of which pair-distances are to be preserved in reduced binary codes as possible, inspired by theory below,
\newtheorem{theorem}{Theorem}
\begin{theorem}\cite{Hyunsoo07DPDR}
The distance $ D\left( x_i, x_j \right) $ in $ L_2 $ norm in the full dimensional space between two vectors $ x_i $ and $ x_j $ is completely preserved in the reduced space obtained from the thin SVD. That is, $ D\left( x_i, x_j \right) = D\left( \widetilde{x_i}, \widetilde{x_j} \right) $, where $ \widetilde{x_i} =  {\widetilde{U}} ^T x_i $ and $ \widetilde{x_j} = {\widetilde{U}} ^T x_j $, and the thin SVD of $ X $ is $ \widetilde{U} \widetilde{\Sigma} {\widetilde {V} }^T  $.
\end{theorem}

Furthermore, there are some other methods focusing on the supervised / semi-supervised extensions of hashing mechanism \cite{Xu11CompleH}, while the training information is utilised to improved the binary coding ability. For example, PCA-like hashing algorithms apply the well-known principle component analysis (PCA) to find the projection functions in an iterative approach \cite{Carreira15HBA}\cite{Heo12sph}\cite{Gong11ITQ}, where most solutions adopt an iteration approach to hash data into a suitable rotation of projections. But they may be in loss of adaptive controlling in such repeated steps, and difficult to handle data stream applications, e.g., online learning \cite{Bach08HMKL}. Furthermore, there are several issues should be highlighted if data stream is conducted for continuous hashing:
\begin{itemize}
 \item As the increasing amount of appetitive set from data stream, the pre-learned hashing may fail to handle new data. And even worse, there could be few available knowledge connecting new hashing codes with previous ones.
 \item For persist and continuous computation, it is unreasonable and almost impossible to explicitly learn hashing function every once coming of new data, while the hashing results of previous data are desired to be absorbed.
\end{itemize}
Though acceptable results can be obtained, iteration is necessary to ensure the efficiency of stepwise improvement. And most of such kind of methods require the calculation of differences among whole data to determine the similarity or unsimilarity between different data. Recently, some novel solutions attempt to tackle the large amounts of data emerge as random hashing \cite{Alexis11RMMH}\cite{Rahimi09RF}, and low calculation of data handling is preferred for further applications.

In this work, such problem is conducted to encode coming data as well as a set of available hashing results, but the pre-defined hashing mechanism is unknown for further handling as illustrated in Fig. 1. Or explained alternatively, the proposed method does not depend on the inherited hashing of labeled data for encoding of new data as no more knowledge could be explicitly learned. Similar to some existing methods, e.g., LSH, the calculation of differences among whole data is unnecessary to find and update nearest neighbours.

\section{Continuous Random Hashing}
Given a data set $ X_0 = \left[ {{x_1},{x_2}, \cdots, {x_{{n_0}}}} \right] \in {\mathds{R}^{d \times {n_0}}} $, while the associated hashing codes may be available at hand, the continuous random hashing (CRH) aims to encode the coming data stream $ {X_1} \in {\mathds{R}^{d \times {n_1}}}, {X_2} \in {\mathds{R}^{d \times {n_2}}}, \cdots, {X_k} \in {\mathds{R}^{d \times {n_k}}} $ in sequence. In the literature, the most matched approaches to the similar problem have been considered as such kind of scalable learning. However, they have proposed the basic hashing procedure with strictly required conditions with supervised or semi-supervised setting, as well as previous labeled information. Though these considerations are able to conduct scalable hashing sharing with the common idea of incremental learning, may fail to make further development on more complicated sources except for concrete assumptions.

The most popular assumption on such scenario is to construct the objective function with knowledge on nearest neighbors (NNs) of each data beforehand, then supervised or semi-supervised handling procedure can be carried out as an optimized hashing with neighborhood information. Distinguishingly, the searching for NNs are not required in CRH to enhance the binary discrimination ability, and further no serve condition is referred for sequential learning of presently coming data. In details, the data sets $ {X_1}, {X_2}, \cdots, {X_{{n_k}}} $ can be afforded by CRH with different sample sizes, and few updates are resorted for handling of each data sets.

Without loss of generic strategy, it is assumed that there is an available mean centered data set $ X \in \mathds{R}^{d \times {p}} $ with binary codes $ A = \left[ {{t_1},{t_2}, \cdots, {t_p}} \right] \in \mathds{R}^{m \times {p}} $ of ideal hashing, where $ m $ usually holds a small size than dimensionality of original data $ d $. Then, the open problem could be described as such that there comes another new data set $ Y = \left[ y_1, y_2, \cdots, y_q \right] \in \mathds{R}^{d \times {q}} \left( {q \ne p} ~is ~possible \right) $, it is required to learn the binary hashing codes $ S = \left[ s_1, s_2, \cdots, s_q \right] \in \mathds{R}^{m \times q} $ of each new data efficiently. As it is designed to preserve data similarity in reduced codes, it is straightforward to optimize the following objective function:
\begin{equation}
 \begin{array}{ll}
Obj(s) & =  \arg \mathop {\min }\limits_{s \in \left\{ { - 1,1} \right\}} \sum\limits_{i = 1}^q {\sum\limits_{j = 1}^p {{{\left\| { H({s_i},{t_j}) - D({y_i},{x_j})} \right\|}_2}} } \\
     & =  \arg \mathop {\min }\limits_{s \in \left\{ { - 1,1} \right\}} \sum\limits_{i = 1}^q {\sum\limits_{j = 1}^p {{{\left\| {\frac{1}{m} D({s_i},{t_j}) - D({y_i},{x_j})} \right\|}_2}} }
\end{array}
\end{equation}
Here, $ {H({\cdot},{\cdot})} $ denotes the hamming distance between two binary codes, affiliated with corresponding Euclidean distance $ D({\cdot},{\cdot}) $ of original data. Nevertheless, it is different to handle the objective with high-order optimization, and almost hardly to be applied if large amounts of data stream are to be sequentially encoded.

To address these limitations, the original data are all normalized $ {x_i}^T{x_i} = 1 $ as done in previous work \cite{Cheng14NCECA}. By defining another function $ g \left( x_i, x_j \right) = {x_i}^T x_j $, then the formalism can be further compressed as
\begin{equation}\label{Eq-1}
\begin{array}{ll}
 Obj\left( s \right) & = \arg \mathop {\min }\limits_{s \in \left\{ { - 1,1} \right\}} \sum\limits_{i = 1}^q {\sum\limits_{j = 1}^p {  {{\left\| { \frac{1}{m} {s_i}^T{t_j} - {y_i}^T{x_j}} \right\|}_2}} } \\
    & = \arg \mathop {\min }\limits_{s \in \left\{ { - 1,1} \right\}} \sum\limits_{i= 1}^q {\sum\limits_{j = 1}^p {  {{\left\| {\frac{1}{m} g\left( {{s_i},{t_j}} \right) - g\left( {{y_i},{x_j}} \right)} \right\|}_2}} }
 \end{array}
\end{equation}
And it could be optionally solved as a standard linear equation. Nevertheless, there are several vital limitations suspended for further applications, e.g.,
\begin{itemize}
\item If large amounts of data are met, system would suffer from ordeal of computational cost on data similarities.
\item The solution requires the inverse of coded data set as a necessary step, which is difficult to get if large data are involved.
\item As the intrinsic differences between original and binary data, hashing codes are hardly to ideally approximate the original data with simple handling, while features are always reduced to short ones.
\end{itemize}

For adaptive considerations, the problem is to be addressed in a random approach to decoding, and is realistically approximate to objective in practice. Here, the following theorem has been referred.
\begin{theorem}\label{Th-1}
For a given data set $ X \in {\mathds{R}^{d \times m}} $ and another data set $ Y \in {\mathds{R}^{d \times n}} $ as described in context, the similarities between them $ G $ can be approximated by rank-$ k $ matrix $ \Pi {U_k} {U_k}^T \Pi $, where $ \Pi $ is formed of $ X $ and $ Y $ as $ \Pi = \left[ X ~ Y\right] $, $ {U_k} $ is the matrix consisting of the left singular vectors of $ C_{\Pi} $. And $ C_{\Pi} $ is a submatrix whose columns consisting of randomly selected from $ \Pi $.  \\
Proof: The proof is given in Appendix.  \\
\end{theorem}
In terms of this consideration, it is feasible to use a subset of original data $ \widehat X $ associated with hashing codes $ \widehat A $ to meet the demand of original problem (\ref{Eq-1}). Thereafter, the calculation could be reduced to a \emph{random} approach with lightweight cost, and piecewise learning is feasible for sequential data.

 More specifically, it is reliable to randomly pick a subset data $ \widehat X \in \mathds{R}^{d \times r} $ from $ X $ to form data set $ \Pi $ with whole coming data $ Y $, and then its decomposition is obtained as
 \begin{equation}
  \Pi  = [\begin{array}{*{20}{c}}
{\widehat X}&Y
\end{array}] = {U_m}{\Lambda _m}{V_m}^T
 \end{equation}
Thus, the inner product of data can be approximated as $ {\Pi^T}{U_m} {U_m}^T \Pi $. And the whole procedure of CRH could be summarized as:
\begin{itemize}
\item Randomly select a subset data set $ \widehat X $ from normalized data set $ X $, and form data $ \Pi  = [\begin{array}{*{20}{c}}
{\widehat X}&Y
\end{array}] $ with coming data $ Y $.
\item  Calculate the low-rank approximate decomposition $ \Pi  = [\begin{array}{*{20}{c}}
{\widehat X}&Y
\end{array}] = {U_m}{\Lambda _m}{V_m}^T $.
\item Solve the optimization problem (\ref{Eq-1}) with approximate data, and obtain the result $ s $.
\end{itemize}

\begin{figure*}\label{Fig-2}
\centering
\subfigure[]{\includegraphics[width=.45\textwidth]{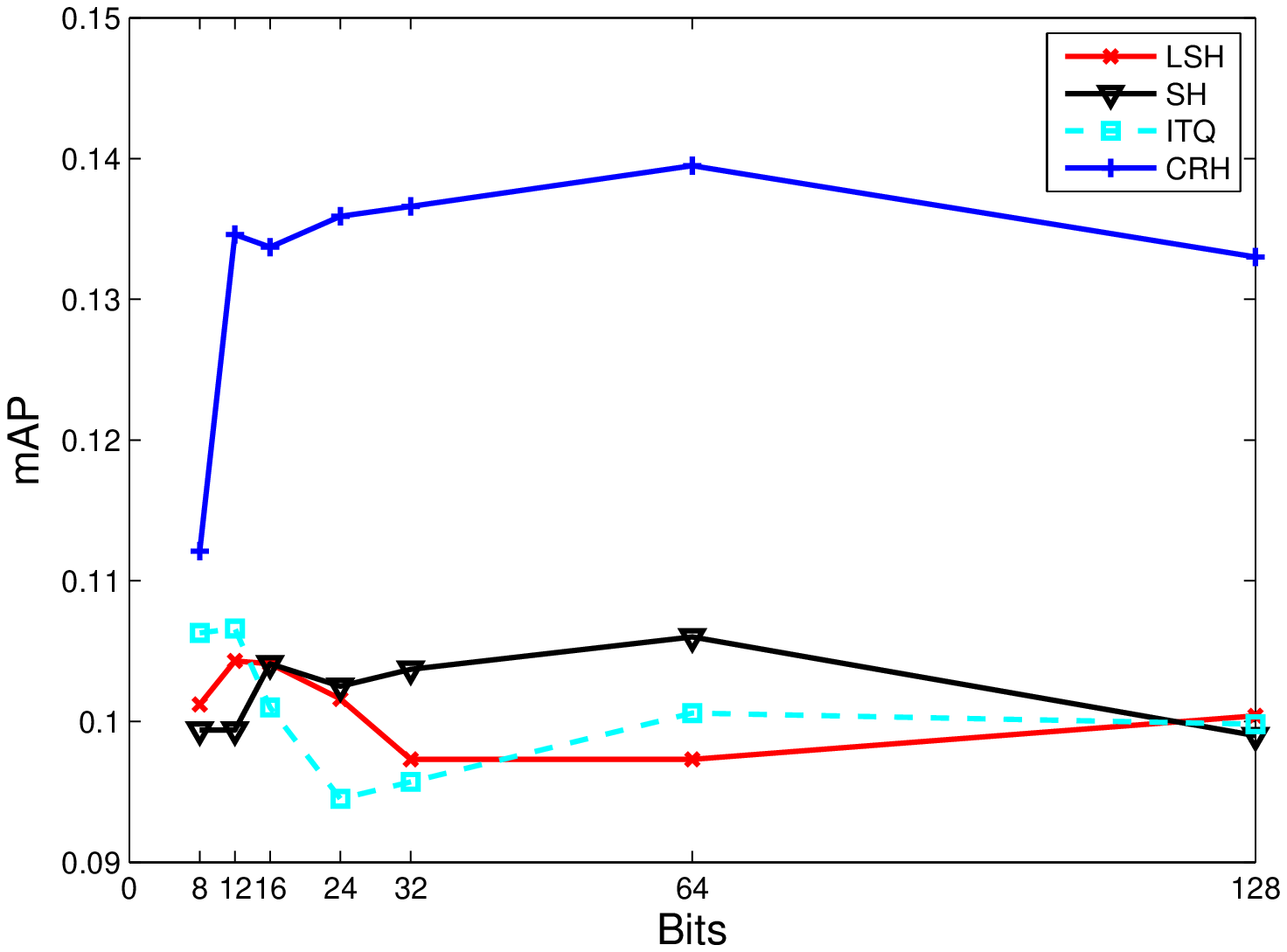}}
\subfigure[]{\includegraphics[width=.45\textwidth]{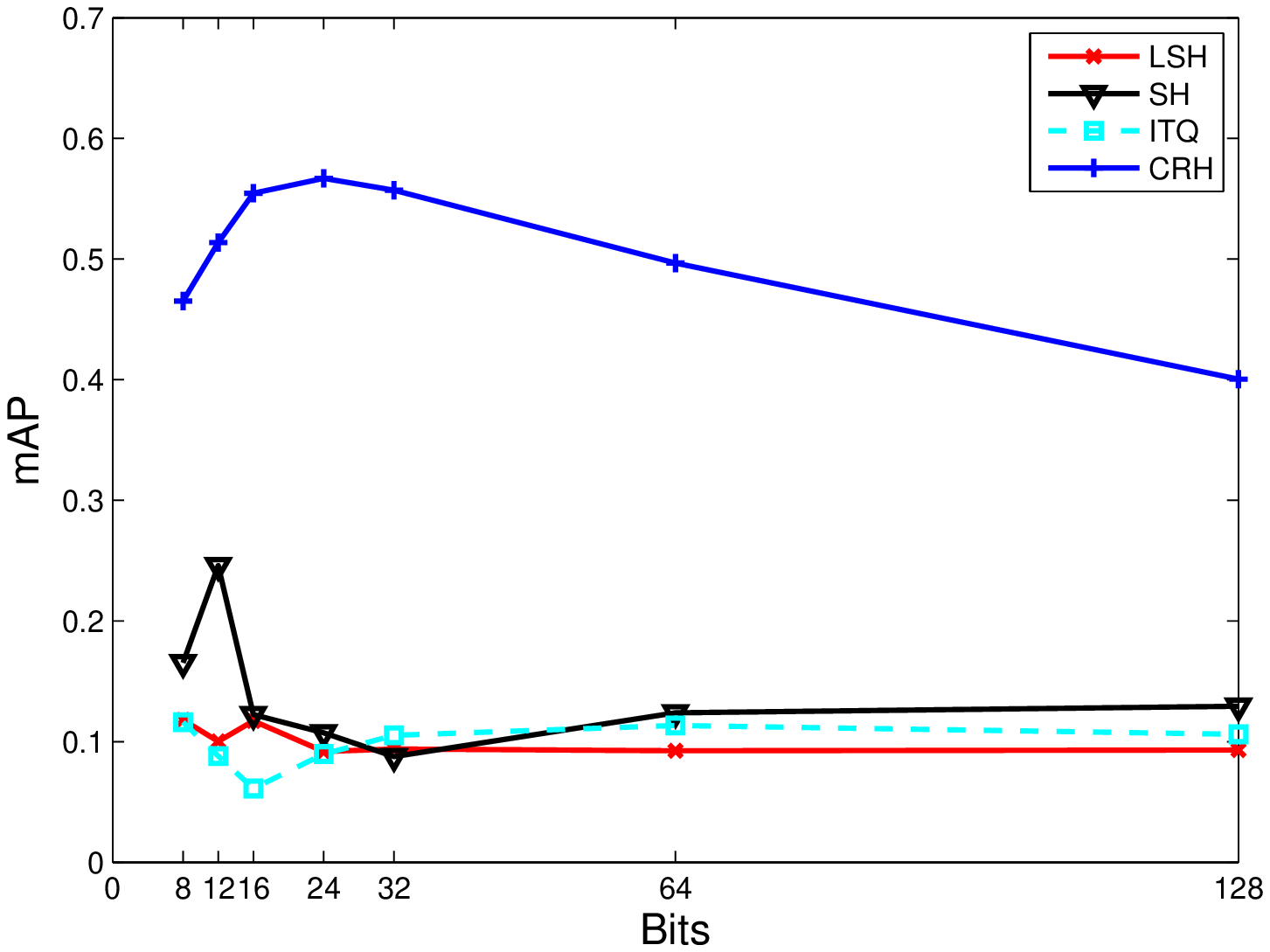}}
\subfigure[]{\includegraphics[width=.45\textwidth]{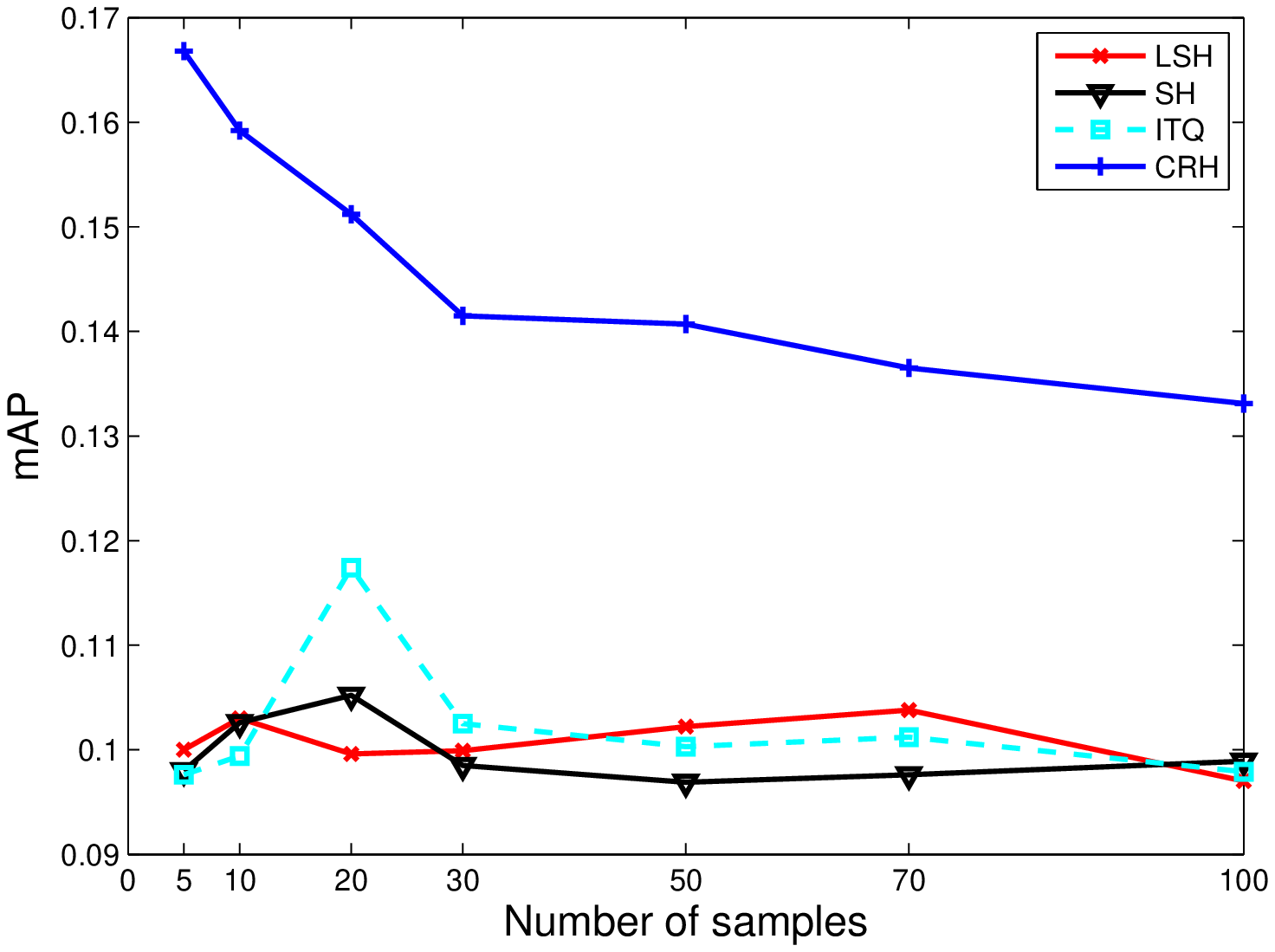}}
\subfigure[]{\includegraphics[width=.45\textwidth]{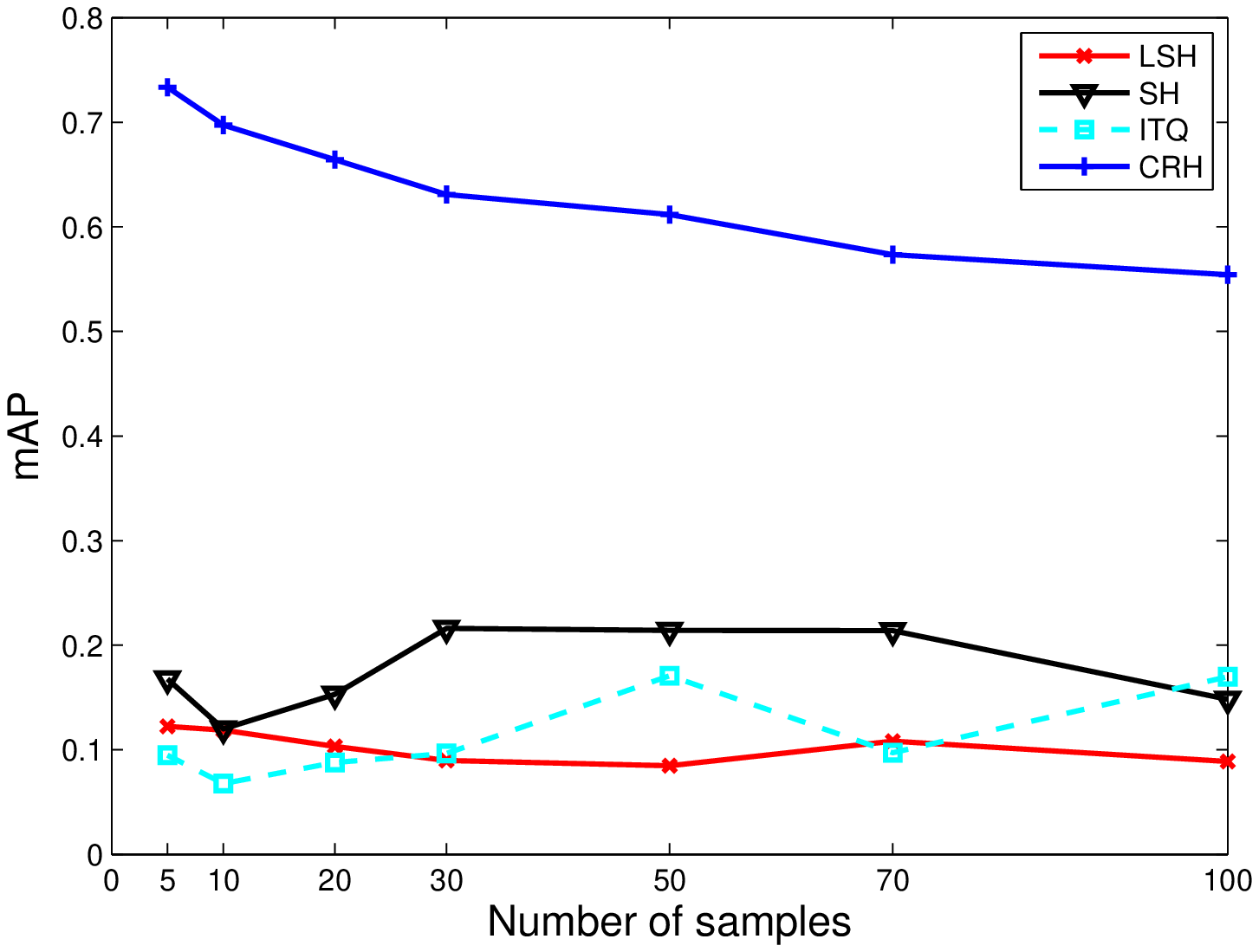}}
\caption{The search results from CIFAR-10 and MNIST datasets. (a) and (b) the
obtained results with changing hashing bits; (c) and (d) the obtained results with different number of involving nearest neighbors in mAP.}
\end{figure*}
\section{Experiments}
In this section, the performance of the proposed method is compared with several benchmark hashing methods, including LSH \cite{LSH08Alexandr}, SH \cite{Weiss09SH}, ITQ \cite{Gong11ITQ}, and RMMH \cite{Alexis11RMMH}. As the continuous hashing is considered, the encoding of testing data is hardly to refer to training stage of previous hashing. As a result, all hashing methods encode query or new data separating from pre-defined hashing functions, and encoders independently learn binary codes of data sets sequentially. Two datasets, CIFAR-10 image dataset \cite{Krizhevsky09cifar} and MNIST digit database \cite{Lecun98minist}, are used to evaluate the hashing results.


For images in CIFAR-10 dataset, a set of 512 dimensional GIST descriptors \cite{Oliva01GIST} and 128 dimensional SIFT descriptors \cite{Lowe04sift} are learned from every tiny image, so that each data is represented as a 640 dimensional sample vector. In the experiments, the different batches in CIFAR-10 dataset is combined into one by ignoring their groups and several subsets are randomly selected to form the aforehand hashing data, query data and sequential data. For each data in MNIST, 784 dimensional gray features are used to describe its visual handwritten patterns. Similarly, the whole data are combined while the original order of data is disordered, and then lots of subsets are randomly selected to be involved into experiments. Furthermore, the performance is measured by the mean average precision (mAP) defining as the average area under the recall-precision curves, and the ground truths are computed based on k nearest neighbors.


In this first experiment, the search performance of encoders are evaluated with two databases. Among all samples of each dataset, 10000 data are randomly selected and encoded with binary hashing, and then 500 data are randomly selected to push forward matching query. For each hashing algorithms, the two data sets are encoded separately, and no supervised / semi-supervised corporation between them is involved. With respect to random column selection of CRH, it is found large quantity of involved columns are to provide convincible performance. However, the selected columns are set to be around 8\% - 10\% of existing data, which is enough to give acceptable performance. The search results with different hashing bits are shown in Fig. 2 (a) and (b), while the obtained results against different number of samples used in mAP matching are shown in Fig. 2 (c) and (d). The experiments are proceeded while fixing the length of hashing bits and involved neighbors in mAP by turns. In details, 100 neighbors are used to measure mAP, and 16 fixed bits are used to generate results with different neighbors.

As illustrated, the results obtained from CRH outperforms other methods at an optimal ability of encoding, the learning binary codes are able to provide distinctive patterns for query. However, it seems proposed random hashing works better on MNIST, for concrete pattern concision. Among other benchmark hashing approaches, SH gives much better performance compared with other unsupervised ones, with robust self-adaptive encoding manner. ITQ also provides acceptable results if enough search information, e.g., hashing bits or nearest neighbors, can be given in searches. This may because it mainly depends on iterative rotation for improvement on projection directions, as an adjustment of learning error. Furthermore, most methods reach the common level of query performance, providing a similar performance of search matching of separate hashing.

\begin{figure}\label{Fig-3}
\centering
\subfigure[]{\includegraphics[width=.54\textwidth]{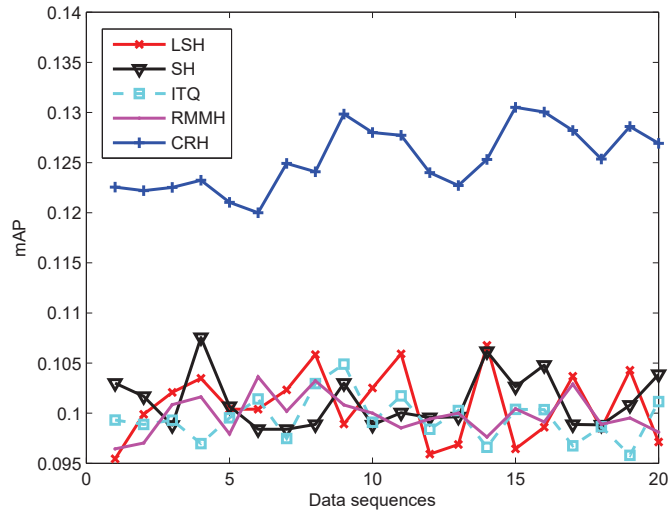}}
\subfigure[]{\includegraphics[width=.54\textwidth]{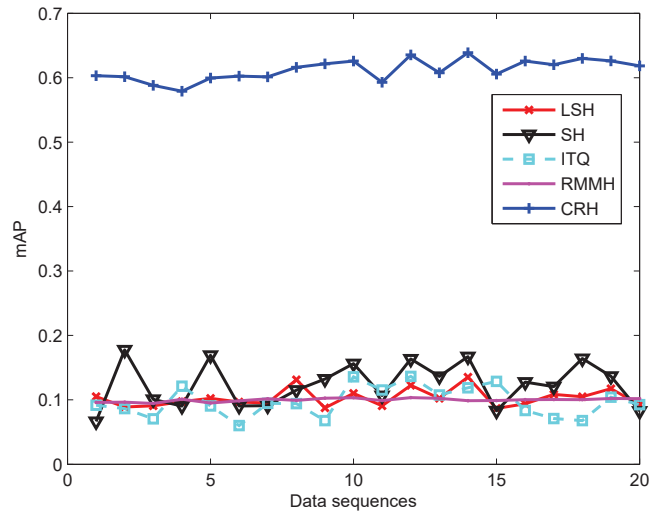}}
\caption{The sequential encoding results from CIFAR-10 and MNIST datasets. (a) Sequential encoding results from CIFAR-10; (b) Sequential encoding results from MNIST.}
\end{figure}
In the second experiment, the continuous hashing ability of different methods are evaluated with a sequence of data stream. At first, 10000 samples are randomly selected from each database, and their binary codes are learned based on different hashing algorithms. Then a series of data sets are absorbed into the encoders by turns, and each data set has 500 data samples for encoding. Another recently proposed random hashing method, namely RMMH \cite{Alexis11RMMH}, is joined for comparison of continuous hashing with randomly selected dispersive data sets.
Noticeably, no supervised / semi-supervised welfare of previously hashing data involve into encoding of currently coming data. Similarly, the binary codes of originally stored data and the appended data are learned separately, while each hashing methods handle the new data as their coming immediately. The encoding results against data stream sequences are shown in Fig. 3 (a) and (b), of which 16 binary bits and 50 NNs are used to measure performance learned from two dispersive groups with 100 randomly selected data.

For continuous hashing of the coming data, most algorithms give the similar results as standard hashing, while the hashing function is suspended for the labeled codes of available data. As the results shown, there is always dynamic fluctuation over decoded matching results as sequential data coming. Though the results are changed over different crests and troughs, the fluctuate area is strictly bounded to attain a stable performance for sequential learning of data stream. Nevertheless, SH is possible to give better results compared with other methods, benefitting from spectral preserving idea. Obviously, CRH is able to overcome the difficulties in sequentially adaptive learning, and provide the useful binary codes with random selected samples.

\section{Conclusions}
As an adaptive encoder learning solution, a continuous hashing method is proposed to learn the binary patterns of  multi-source data. The motivation is based on the fact that encoder may be unavailable for binary hashing of further sequential data, e.g., data stream. This pushes forward much attentions on designing efficient solutions to continuous hashing. Different from existing methods, the proposed benchmark approach is able to learn the binary codes without previously hashing knowledge involved, and iterative optimization can be avoided in an efficient manner. Furthermore, adaptive hashing is component to continuous learning by benefitting from a random selection of coded data. As experimental results shown, the proposed method is able to provide outstanding performance for sequential hashing compared with other benchmark methods.

\section{Proof of Theorem 2}
\newtheorem{lemma}{Lemma}
\begin{lemma}\cite{Drineas05GM}
Given the matrix $ G \in \mathds{R}^{n \times n} $ with $ G_{ij} = g(x_i, x_j) $ and $ {p_i} = G_{ii}^2/\sum\nolimits_{i = 1}^n {G_{ii}^2} \left( {\sum\nolimits_{i = 1}^n {{p_i}}  = 1} \right) $, the original matrix could be approximated as $ \widetilde G = C {W_k}^ + {C^T} $, where $ C $ is the selected columns of $ G $ according to the sampling probabilities $ p_i $, and $ W_k $ is the best rank-$ k $ approximation to $ W $, the matrix formed by the intersection between those $ c $ columns of $ G $ and the corresponding $ c $ rows of $ G $.
\end{lemma}
Proof: Following the above lemma, the original matrix is able to be approximated by randomly selected $ C $ from $ G $ with uniform probability. By introducing the sampling matrix $ S \in {\mathds{R}^{n \times c}} $ as a zero-one matrix where $ S_{ij} = 1 $ if the $ i $-th column of $ X $ is chosen and $ S_{ij} = 0 $ otherwise, the selected columns can be described as $ C = GSD \in {\mathds{R}^{d \times c}} $ and $ D $ is the rescaling diagonal matrix $ D \in \mathds{R}^{c \times c} $ with $ {D_{tt}} = {1 \mathord{\left/
 {\vphantom {1 {\sqrt {c{p_{{i_t}}}} }}} \right.
 \kern-\nulldelimiterspace} {\sqrt {c{p_{{i_t}}}} }} $. Then, it is clear that $ C_\Pi = \Pi SD $ and $ W = {\left( SD \right)}^T GSD $. Let the
 rank-$ k $ decomposition of $ C_\Pi $ be $ {C_\Pi} = {\widetilde U_k}{\widetilde \Sigma _k} {\widetilde {V_k} }^T $, the approximate matrix can be calculated as $ {\Pi ^T}{\widetilde U_k}\widetilde {U_k}^T \Pi $.

\section*{Acknowledgment}
This work was supported by research committee of Macau University of Science and Technology.
The corresponding author of this work is Dr. Miao Cheng.



%
%

\bibliographystyle{IEEEtran}
\bibliography{IEEEabrv,CRHRef}

\end{document}